\documentclass[10pt,a4paper]{article}

\usepackage[utf8]{inputenc}
\usepackage[T1,T5]{fontenc}
\usepackage[a4paper,margin=1in]{geometry}

\usepackage{booktabs}
\usepackage{amsmath}
\usepackage{pgfplots}
\pgfplotsset{compat=1.18}
\usepackage{graphicx}
\usetikzlibrary{patterns,arrows.meta,calc}
\definecolor{lost}{RGB}{160,32,26}
\definecolor{tr}{RGB}{29,66,138}
\usepackage{url}
\usepackage[final]{microtype}
\usepackage[hidelinks]{hyperref}

\newcommand{\keywords}[1]{%
  \begingroup\vspace{0.4em}\noindent\textbf{Keywords:} #1\par\vspace{0.4em}\endgroup}

\begin{document}

\title{Measuring the Partial-Credit Gap: A Strict Benchmark on Vietnam's 2025 Convex Marking Scheme}

\author{Nguyen Quoc Hung\textsuperscript{*}, Nguyen Dang Minh, Le Nhu Quynh, \\
Tran Khanh Linh, Nguyen Kieu Linh \\
Posts and Telecommunications Institute of Technology, Hanoi, Vietnam \\
nguyenquochung.workvn@gmail.com \\
\textsuperscript{*}Corresponding author}
\date{}

\maketitle

\begin{abstract}
When evaluating language models on human exams, benchmarks typically score each response as right or wrong and report the overall accuracy. This approach assumes that partial knowledge is worth proportional credit, an assumption that fails when an examination uses a non-additive grading scheme. The 2025 reform of Vietnam's National High School Graduation Examination demonstrates the cost of this substitution. In Part~II of the exam, candidates evaluate four true/false statements per question. The grading is convex: the number of correct statements earns 0, 0.10, 0.25, 0.50, or 1.00 points. Identifying three statements correctly pays 0.50 points, not the 0.75 points that standard accuracy metrics would award. Because Part~II accounts for 4.00 of the exam's 10.00 points, reporting accuracy inflates the score by rewarding partial knowledge that the state explicitly penalizes. We introduce THPT-Ladder, a benchmark of 632 items from 21 official exams across 11 subjects, graded exactly as the ministry grades its students. The ministry publishes the marks of over a million candidates, allowing us to place models directly into the human cohort. Across eight models, the official rubric pays 0.020 to 0.159 points less per Part~II question than proportional credit. This shortfall changes a model's apparent competence. For Qwen3.5-27B on the 2025 History exam, a 0.042-point shortfall drops its standing from the 90th to the 77th percentile among 481{,}293 candidates. A model's accuracy does not predict this penalty. At Claude Sonnet 5's accuracy level, different distributions of errors yield scores varying from 0.869 to 0.932 points per question. Official marks depend on how correct statements are grouped, meaning standard benchmarks report a competence the institution would not certify.
\end{abstract}


\section{Introduction} \label{sec:intro}

The evaluation of language models heavily relies on examinations written for human candidates, ranging from multitask suites assembled out of school exams~\cite{indommlu,arabicmmlu} to single-country benchmarks~\cite{vmlu,vnhsge}. However, almost all of these benchmarks keep the questions but discard the marking scheme. They score each item right or wrong and report accuracy. This substitution assumes that partial knowledge is always worth proportional credit. In reality, an examination's marking scheme encodes exactly how much partial knowledge is worth. Replacing that scheme with flat accuracy credits a model for every fragment it answers correctly, allowing it to appear highly competent on an exam it would have actually failed.

\begin{figure}[t]\centering
\resizebox{0.82\linewidth}{!}{
\begin{tikzpicture}[x=1cm,y=1cm,font=\scriptsize]
\fill[black!12] (0.000,0) -- (0.000,0.000) -- (0.182,0.000) -- (0.365,0.000) -- (0.547,0.000) -- (0.730,0.002) -- (0.912,0.006) -- (1.095,0.017) -- (1.277,0.048) -- (1.460,0.112) -- (1.643,0.229) -- (1.825,0.381) -- (2.008,0.600) -- (2.190,0.844) -- (2.373,1.136) -- (2.555,1.439) -- (2.737,1.741) -- (2.920,1.963) -- (3.103,2.194) -- (3.285,2.405) -- (3.467,2.521) -- (3.650,2.579) -- (3.833,2.620) -- (4.015,2.597) -- (4.197,2.498) -- (4.380,2.307) -- (4.562,2.111) -- (4.745,1.847) -- (4.928,1.559) -- (5.110,1.263) -- (5.292,0.981) -- (5.475,0.728) -- (5.657,0.498) -- (5.840,0.345) -- (6.022,0.212) -- (6.205,0.132) -- (6.388,0.068) -- (6.570,0.034) -- (6.753,0.017) -- (6.935,0.009) -- (7.117,0.003) -- (7.300,0.001) -- (7.300,0.000) -- (7.300,0) -- cycle;
\draw[black!45,line width=.5pt] (0.000,0.000) -- (0.182,0.000) -- (0.365,0.000) -- (0.547,0.000) -- (0.730,0.002) -- (0.912,0.006) -- (1.095,0.017) -- (1.277,0.048) -- (1.460,0.112) -- (1.643,0.229) -- (1.825,0.381) -- (2.008,0.600) -- (2.190,0.844) -- (2.373,1.136) -- (2.555,1.439) -- (2.737,1.741) -- (2.920,1.963) -- (3.103,2.194) -- (3.285,2.405) -- (3.467,2.521) -- (3.650,2.579) -- (3.833,2.620) -- (4.015,2.597) -- (4.197,2.498) -- (4.380,2.307) -- (4.562,2.111) -- (4.745,1.847) -- (4.928,1.559) -- (5.110,1.263) -- (5.292,0.981) -- (5.475,0.728) -- (5.657,0.498) -- (5.840,0.345) -- (6.022,0.212) -- (6.205,0.132) -- (6.388,0.068) -- (6.570,0.034) -- (6.753,0.017) -- (6.935,0.009) -- (7.117,0.003) -- (7.300,0.001) -- (7.300,0.000);
\draw[black!70,line width=.7pt] (0,0) -- (7.30,0);
\draw[black!70,line width=.5pt] (0.000,0) -- (0.000,-0.09);
\node[below,inner sep=1.5pt,black!60] at (0.000,-0.09) {0};
\draw[black!70,line width=.5pt] (1.460,0) -- (1.460,-0.09);
\node[below,inner sep=1.5pt,black!60] at (1.460,-0.09) {2};
\draw[black!70,line width=.5pt] (2.920,0) -- (2.920,-0.09);
\node[below,inner sep=1.5pt,black!60] at (2.920,-0.09) {4};
\draw[black!70,line width=.5pt] (4.380,0) -- (4.380,-0.09);
\node[below,inner sep=1.5pt,black!60] at (4.380,-0.09) {6};
\draw[black!70,line width=.5pt] (5.840,0) -- (5.840,-0.09);
\node[below,inner sep=1.5pt,black!60] at (5.840,-0.09) {8};
\draw[black!70,line width=.5pt] (7.300,0) -- (7.300,-0.09);
\node[below,inner sep=1.5pt,black!60] at (7.300,-0.09) {10};
\node[below,inner sep=1pt,black!60] at (3.65,-0.42) {mark out of 10.00 (282,519 candidates)};
\draw[black!55,dashed,line width=.5pt] (3.650,0) -- (3.650,2.68);
\node[black!55,anchor=south,inner sep=1.5pt] at (3.650,2.68) {pass 5.00};
\draw[lost,line width=1.0pt] (1.277,0) -- (1.277,0.048);
\filldraw[lost] (1.277,0.048) circle (0.045);
\node[lost,anchor=south,inner sep=1.5pt,align=center] at (1.277,0.108) {\texttt{SDSS}\\1.75};
\draw[black,line width=1.0pt] (4.745,0) -- (4.745,1.847);
\filldraw[black] (4.745,1.847) circle (0.045);
\draw[black,line width=1.0pt] (5.110,0) -- (5.110,1.263);
\filldraw[black] (5.110,1.263) circle (0.045);
\draw[black,line width=1.0pt] (6.022,0) -- (6.022,0.212);
\filldraw[black] (6.022,0.212) circle (0.045);
\draw[black,line width=1.0pt] (6.022,0) -- (6.022,0.212);
\filldraw[black] (6.022,0.212) circle (0.045);
\draw[black,line width=1.0pt] (6.388,0) -- (6.388,0.068);
\filldraw[black] (6.388,0.068) circle (0.045);
\draw[black,line width=1.0pt] (6.388,0) -- (6.388,0.068);
\filldraw[black] (6.388,0.068) circle (0.045);
\draw[black,line width=1.0pt] (6.753,0) -- (6.753,0.017);
\filldraw[black] (6.753,0.017) circle (0.045);
\draw[black,line width=1.0pt] (6.935,0) -- (6.935,0.009);
\filldraw[black] (6.935,0.009) circle (0.045);
\node[anchor=south west,inner sep=1.5pt,align=left] at (6.635,0.069) {8 models\\6.50--9.50};
\end{tikzpicture}}
\caption{Marks of the 282{,}519 candidates who sat Economics \& Law in 2026, with all eight models and a fixed answer string placed on the same scale. Each marker is drawn at its mark, with height proportional to the number of candidates who obtained that mark. On this exam the best fixed string is \texttt{SDSS}, which earns 1.75 without reading any question. Scoring under the official rules locates a model within the population the examination was written to assess, an outcome that standard accuracy cannot achieve.}
\label{fig:cohort}
\end{figure}
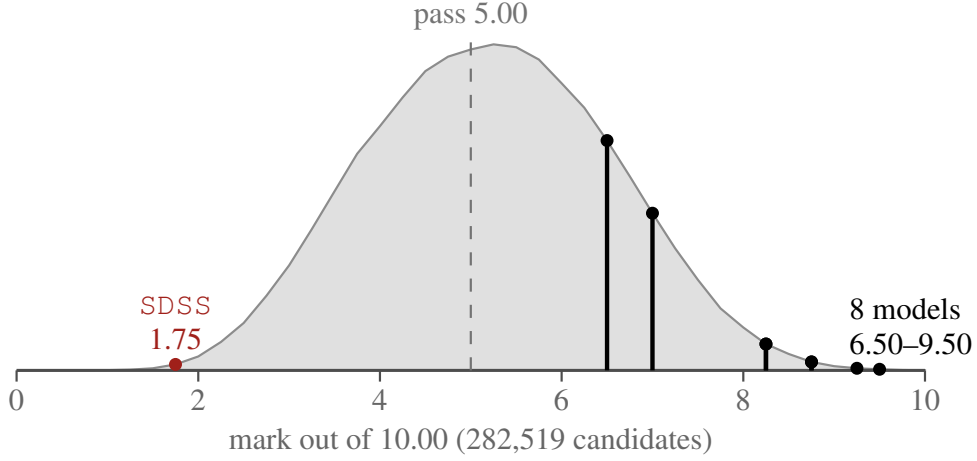

Vietnam's National High School Graduation Examination (\emph{Kỳ thi tốt nghiệp trung học phổ thông}, THPT) makes the cost of ignoring the marking scheme measurable. The examination was sat by 1.13 million candidates in 2025. The Ministry of Education and Training publishes the exams, the answer keys, and the binding rules that dictate exactly how each format is marked.

These rules changed in 2025. The ministry restructured the examination, and Decision~764/QĐ-BGDĐT~\cite{qd764} now defines three question formats: a four-option item worth 0.25 points, four true/false statements marked together as one question, and a short-answer item with no options. The exams print these as \emph{PHẦN I}, \emph{PHẦN II} and, where a subject carries it, \emph{PHẦN III}, and we refer to them as Part~I, Part~II and Part~III. No existing Vietnamese benchmark covers the two new formats. The largest existing dataset, VNHSGE, draws on exams from 2019 to 2023, predating the reform entirely.

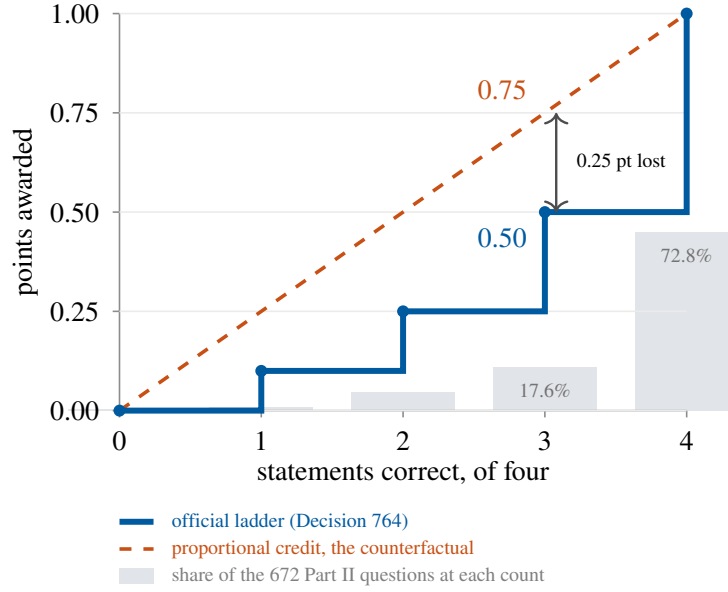
\begin{figure}[t]\centering
\resizebox{0.62\linewidth}{!}{
\begin{tikzpicture}[x=1cm,y=1cm]
\footnotesize
\definecolor{ladder}{RGB}{0,90,160}
\definecolor{propc}{RGB}{200,80,20}
\definecolor{distbar}{RGB}{225,228,232}
\fill[distbar] (-0.455,0) rectangle (0.455,0.010);
\fill[distbar] (0.795,0) rectangle (1.705,0.032);
\fill[distbar] (2.045,0) rectangle (2.955,0.167);
\fill[distbar] (3.295,0) rectangle (4.205,0.380);
\node[font=\tiny,text=black!55,anchor=north] at (3.750,0.340) {17.6\%};
\fill[distbar] (4.545,0) rectangle (5.455,1.575);
\node[font=\tiny,text=black!55,anchor=north] at (5.000,1.535) {72.8\%};
\draw[black!45] (0,0) -- (5.000,0);
\draw[black!45] (0,0) -- (0,3.500);
\draw[black!45] (0.000,0) -- (0.000,-0.06);
\node[font=\scriptsize,anchor=north] at (0.000,-0.08) {0};
\draw[black!45] (1.250,0) -- (1.250,-0.06);
\node[font=\scriptsize,anchor=north] at (1.250,-0.08) {1};
\draw[black!45] (2.500,0) -- (2.500,-0.06);
\node[font=\scriptsize,anchor=north] at (2.500,-0.08) {2};
\draw[black!45] (3.750,0) -- (3.750,-0.06);
\node[font=\scriptsize,anchor=north] at (3.750,-0.08) {3};
\draw[black!45] (5.000,0) -- (5.000,-0.06);
\node[font=\scriptsize,anchor=north] at (5.000,-0.08) {4};
\draw[black!45] (0,0.000) -- (-0.06,0.000);
\node[font=\scriptsize,anchor=east] at (-0.08,0.000) {0.00};
\draw[black!8] (0,0.000) -- (5.000,0.000);
\draw[black!45] (0,0.875) -- (-0.06,0.875);
\node[font=\scriptsize,anchor=east] at (-0.08,0.875) {0.25};
\draw[black!8] (0,0.875) -- (5.000,0.875);
\draw[black!45] (0,1.750) -- (-0.06,1.750);
\node[font=\scriptsize,anchor=east] at (-0.08,1.750) {0.50};
\draw[black!8] (0,1.750) -- (5.000,1.750);
\draw[black!45] (0,2.625) -- (-0.06,2.625);
\node[font=\scriptsize,anchor=east] at (-0.08,2.625) {0.75};
\draw[black!8] (0,2.625) -- (5.000,2.625);
\draw[black!45] (0,3.500) -- (-0.06,3.500);
\node[font=\scriptsize,anchor=east] at (-0.08,3.500) {1.00};
\draw[black!8] (0,3.500) -- (5.000,3.500);
\draw[propc,dashed,line width=0.9pt] (0,0) -- (5.000,3.500);
\draw[ladder,line width=1.5pt] (0.000,0.000) -- (1.250,0.000) -- (1.250,0.350);
\draw[ladder,line width=1.5pt] (1.250,0.350) -- (2.500,0.350) -- (2.500,0.875);
\draw[ladder,line width=1.5pt] (2.500,0.875) -- (3.750,0.875) -- (3.750,1.750);
\draw[ladder,line width=1.5pt] (3.750,1.750) -- (5.000,1.750) -- (5.000,3.500);
\fill[ladder] (0.000,0.000) circle (1.5pt);
\fill[ladder] (1.250,0.350) circle (1.5pt);
\fill[ladder] (2.500,0.875) circle (1.5pt);
\fill[ladder] (3.750,1.750) circle (1.5pt);
\fill[ladder] (5.000,3.500) circle (1.5pt);
\draw[black!70,line width=0.6pt,<->] (3.850,1.750) -- (3.850,2.625);
\node[font=\tiny,anchor=west,align=left] at (3.930,2.188) {0.25 pt lost};
\node[font=\scriptsize,text=ladder,anchor=north east] at (3.690,1.710) {0.50};
\node[font=\scriptsize,text=propc,anchor=south east] at (3.690,2.645) {0.75};
\node[font=\scriptsize,anchor=north] at (2.500,-0.34) {statements correct, of four};
\node[font=\scriptsize,rotate=90,anchor=south] at (-0.62,1.750) {points awarded};
\draw[ladder,line width=1.5pt] (0,-0.980) -- (0.30,-0.980);
\node[font=\tiny,text=ladder,anchor=west] at (0.36,-0.980) {official ladder (Decision 764)};
\draw[propc,dashed,line width=0.9pt] (0,-1.220) -- (0.30,-1.220);
\node[font=\tiny,text=propc,anchor=west] at (0.36,-1.220) {proportional credit, the counterfactual};
\fill[distbar] (0,-1.530) rectangle (0.30,-1.390);
\node[font=\tiny,text=black!50,anchor=west] at (0.36,-1.460) {share of the 672 Part~II questions at each count};
\end{tikzpicture}}
\caption{The marking rule and where the answers actually fall. The state pays the step function, while proportional credit would pay the straight diagonal line. The gap is widest when three out of four statements are correct, paying 0.50 points instead of the implied 0.75 points. Notably, 17.6\% of the Part~II questions our models answered landed exactly on this rung. The shortfall in Eq.~\ref{eq:pcg} represents the vertical distance between the two lines, weighted by how often each count occurs. Because the mass of answers sits where the gap is widest, the penalty is a real effect rather than a theoretical edge case.}
\label{fig:ladder}
\end{figure}

Part~II makes the value of partial knowledge visible because it is the only format that uses a non-additive grading scheme. The number of a question's four statements judged correctly indexes a convex step function $\mathrm{ladder}=(0, 0.10, 0.25, 0.50, 1.00)$. Getting three out of four statements correct pays 0.50 points rather than proportional 0.75 points (Fig.~\ref{fig:ladder}). Where this format appears, it carries 4.00 of an exam's 10.00 points. Additive credit would pay $0.25a_q$ where $a_q$ of the four statements are judged correctly, while the state pays $\mathrm{ladder}[a_q]$. We define this difference as the \emph{partial-credit gap}, or \emph{shortfall}. This gap measures the shape of a model's knowledge. A model that is broadly but imperfectly right will fall into the punished zone, while a model that either sweeps a question entirely or fails it completely will not. The shortfall is therefore a calibration signal, not just a second accuracy metric (Section~\ref{sec:metrics}).

Reporting this shortfall requires a baseline, and the standard approaches fail. Marking the four statements additively produces a score the ministry would never award and erases the shortfall by construction. Alternatively, calling 0.25 the chance level ignores what a convex scheme does to a guesser. Flipping a coin for every statement has an expected value of
\begin{equation}
\textstyle\sum_{k=0}^{4}\binom{4}{k}2^{-4}\,\mathrm{ladder}[k] = \tfrac{4.90}{16} = 0.30625,
\label{eq:random}
\end{equation}
compared to 0.25 for a traditional four-option item. The ladder does suppress lucky full marks, as all four statements must land correctly, a 1 in 16 chance. However, it actually raises the guesser's expected payout by 22.5\% above 0.25. If we propagate Eq.~\ref{eq:random} through each subject's published structure, the whole-exam random baseline varies. It runs from 19.75\% of the total scale in Mathematics to 27.25\% in the five subjects that contain 24 Part~I items and no Part~III. A single ``better than chance'' threshold therefore misrepresents both the value of random guessing and the varying structure of the exams (Section~\ref{sec:exam}).

We resolve these errors by applying the ministry's own published rules. We introduce THPT-Ladder, named after the school level \emph{trung học phổ thông}. The corpus contains 632 items from 21 exams across 11 subjects, with every answer drawn directly from the ministry's official key (Section~\ref{sec:dataset}). Because an exam's difficulty fluctuates between years, we report every score as a percentile among the candidates who sat that exact exam (Fig.~\ref{fig:cohort}, Section~\ref{sec:metrics}). This cohort is the population the examination was designed for. Since the state publishes the score distribution, we can interpret a model's mark on the very scale the examination uses.

We evaluated eight models under this rule to measure the substitution cost. The test included three open-weight models and five closed models from two vendors. Qwen3.5-27B judged 92.6\% of Part~II statements correctly but completed only 81.0\% of Part~II questions. The ladder therefore paid it 0.884 points per question, whereas statement-by-statement credit would pay 0.926. Standard benchmarks report the higher of these two figures, while the ministry awards the lower. By using the published cohort marks, we can read this difference as a drop in standing rather than just a decimal. On the 2025 History exam, this penalty drops the model from the 90th percentile to the 77th. The shortfall runs from 0.020 to 0.159 points per question. It is tempting to read this ordering as the ladder heavily taxing knowledge that is broad but scattered.

The closed models show that this interpretation is too simple. When ordered by statement accuracy, the shortfall is not monotone. Accuracy simply does not determine what a non-additive scheme pays (Section~\ref{sec:experiments}). This effect is not specific to any particular model. The shortfall follows mathematically from the ladder, and the guessing floor follows from how the keys are constructed. Any system scored under Decision~764 will encounter both.

We make four contributions. First, we release THPT-Ladder, a benchmark of 632 items from 21 official exams containing every ministry key and marking rule, allowing models to be scored as candidates rather than by flat accuracy. Second, we formalize the shortfall, which measures how far a convex grading scheme's award falls below the accuracy a benchmark would normally report, and provide cohort distributions to convert marks into human percentiles. Third, we show that the published keys are not statement-balanced. A fixed answer string can exploit this imbalance to earn 11.07\% to 24.25\% of an exam without reading a single question. Finally, we release the extraction pipeline used to build the corpus. This includes figure extraction, per-variant key reading, and parsers for the three formats introduced by Decision~764, ensuring that future exams can be added with minimal effort.

\section{Related Work}

Two Vietnamese benchmarks are closely related to this work, but neither can evaluate models under a non-additive grading scheme. VMLU~\cite{vmlu} measures subject knowledge across 58 subjects and four levels of education. It draws on assorted material rather than a single examination, and its multiple-choice component grades items strictly as right or wrong. VNHSGE~\cite{vnhsge} is built directly from the high school graduation exam, featuring over 19{,}000 multiple-choice items and 300 literary essays. However, these materials come from the curriculum that preceded the 2025 reform, and the benchmark evaluates two closed chat services rather than any open-weight models. Because both benchmarks describe the exam as it stood before 2025 and lack a marking scheme, neither can distinguish a model that gets three out of four statements correct from one that gets all four correct. Vietnam is not unusual in this respect. Curriculum-aligned benchmarks for other lower-resource languages, including LaoBench~\cite{laobench}, SinhalaMMLU~\cite{sinhalammlu}, and an assessment against Nepal's K-10 curriculum~\cite{nepalk10}, also score items as purely right or wrong. We are not aware of any public dataset that includes items in the new 2025 formats.

Scoring models at a finer grain than the whole item is established practice in other domains. SteuerLLM~\cite{steuerllm} is the closest to our setting, marking German tax-law questions statement by statement. RadSEM~\cite{radsem} decomposes radiology reports into atomic findings and refuses credit when errors exist. PsyScore~\cite{psyscore} applies a graded partial-credit item-response model, and CMPhysBench~\cite{cmphysbench} awards non-binary credit over expression trees. However, none of these benchmarks uses a convex credit function where three statements out of four pay 0.50 instead of the proportional 0.75. We are not aware of any prior benchmark that scores language models against a non-additive grading scheme established by a national ministry, let alone one that evaluates the models against the actual human population that took the exam.

Researchers usually test option-order sensitivity by synthetically permuting choices~\cite{ordersens}. The Vietnamese examination instead distributes 24 or 48 official variants (\emph{mã đề}) of every exam. These variants are not just reordered copies of each other. Out of the 20 subject-years where we hold the keys, 15 split into groups with entirely different content. We release every published key to support future research into these variations.

\section{The Examination and the Corpus} \label{sec:exam}\label{sec:dataset}

The examination's published structure fixes both the shortfall and the answer-only baseline. Every candidate sits four subjects. Each exam is marked out of 10.00 and, since the 2025 reform, follows the three question formats defined by Decision~764. The ministry sets the marking scheme, and every answer in our corpus comes directly from the ministry's published keys.

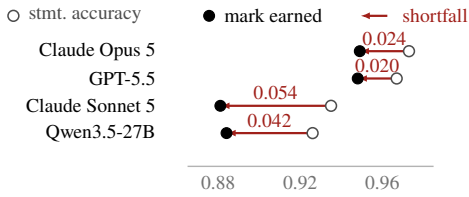
\begin{figure}[t]\centering
\resizebox{\linewidth}{!}{
\begin{tikzpicture}[x=1cm,y=1cm,font=\footnotesize]
\def\LW{7.30}\def\RW{7.30}\def\TOP{0}

\fill[black!8] (0,{\TOP}) rectangle ({\LW+\RW+0.40},{\TOP+0.62});
\draw[black!75,line width=.9pt] (0,{\TOP}) -- ({\LW+\RW+0.40},{\TOP});
\draw[black!75,line width=.9pt] (0,{\TOP+0.62}) -- ({\LW+\RW+0.40},{\TOP+0.62});
\node[anchor=west,inner sep=6pt] at (0,{\TOP+0.31})
  {Biology, upper-secondary graduation examination in Vietnam, 2025 --- PHẦN II (Part~II)};

\def\BODY{-3.28}
\node[anchor=north west,text width={(\LW-0.30)*1cm},align=left,inner sep=6pt] (vi) at (0,{\TOP-0.06})
  {\textbf{Câu 3.} Hình bên thể hiện sự di truyền của 2 tính trạng bao gồm hội chứng
   nail-patella và hệ nhóm máu ABO ở một gia đình. \dots{} Hai gene (N, I) cùng nằm trên
   NST số 9 và có tần số hoán vị gene là 10\%.\\[2pt]
   \textbf{a)} Quần thể người có tối đa 8 kiểu hình liên quan đến 2 tính trạng này.\\
   \textbf{b)} Người II-1 tạo ra 2 loại giao tử mang gene quy định về 2 tính trạng này.\\
   \textbf{c)} Kiểu hình của người I-3 được quy định bởi 1 trong 5 loại kiểu gene về 2
   tính trạng này.\\
   \textbf{d)} Cặp vợ chồng II-1 và II-2 sinh con đầu lòng, xác suất để người con này
   mắc hội chứng nail-patella và có nhóm máu AB là 22,5\%.};
\node[anchor=north west,text width={(\RW-0.30)*1cm},align=left,inner sep=6pt,text=tr]
  (en) at ({\LW+0.30},{\TOP-0.06})
  {\textbf{Question 3.} The figure shows the inheritance of two traits, nail-patella
   syndrome and the ABO blood group, in one family. \dots{} The two genes lie on
   chromosome 9 with a recombination frequency of 10\%.\\[2pt]
   \textbf{a) True.} At most 8 phenotypes for the two traits exist in the population.\\
   \textbf{b) True.} Individual II-1 produces 2 gamete types for these two traits.\\
   \textbf{c) True.} The phenotype of I-3 arises from 1 of 5 genotypes.\\
   \textbf{d) True.} For the first child of II-1 and II-2, P(nail-patella and blood
   group AB) $=$ 22.5\%.};
\path let \p1=(vi.south), \p2=(en.south) in
  coordinate (btm) at (0,{min(\y1,\y2)-2mm});
\draw[black!75,dashed,line width=.8pt] ({\LW+0.20},{\TOP}) -- ({\LW+0.20},0|-btm);
\draw[black!75,line width=.9pt] (0,0|-btm) -- ({\LW+\RW+0.40},0|-btm);

\coordinate (ktop) at (0,0|-btm);
\coordinate (kbot) at ($(ktop)+(0,-0.10)$);
\begin{scope}[shift={($(kbot)+(0.62,-2.02)$)}]
  \def\RS{0.62}\def\MS{0.52}\def\MX{2.62}
  \node[anchor=west,font=\footnotesize\bfseries,inner sep=0pt] at (-0.60,{2.30*\RS})
    {What each model answered};
  \foreach \j/\lt in {0/{a},1/{b},2/{c},3/{d}}{
    \node[font=\scriptsize,black!55] at ({\MX+\j*\MS},{1.74*\RS}) {(\lt)};}
  \node[anchor=east,font=\scriptsize,black!70,inner sep=3pt] at ({\MX-0.5*\MS},{1.74*\RS})
    {key};
  \foreach \j in {0,1,2,3}{
    \node[font=\footnotesize\bfseries] at ({\MX+\j*\MS},{1.20*\RS}) {Đ};}
  \node[anchor=east,font=\scriptsize,black!70,inner sep=3pt] at ({\MX-0.5*\MS},{1.20*\RS})
    {\ };
  \foreach \i/\nm/\ra/\rb/\rc/\rd/\ok/\pay in {%
      0/{Claude Opus 5}/{Đ}/{Đ}/{Đ}/{Đ}/4/{1.00},
      1/{InternVL3.5-8B}/{Đ}/{S}/{Đ}/{Đ}/3/{0.50},
      2/{Claude Haiku 4.5}/{Đ}/{S}/{Đ}/{S}/2/{0.25},
      3/{Qwen3.5-9B}/{S}/{S}/{S}/{Đ}/1/{0.10},
      4/{Qwen3.5-27B}/{--}/{--}/{--}/{--}/0/{0.00}}{
    \pgfmathsetmacro\yy{(0.86-\i*0.52)*\RS}
    \node[anchor=east,font=\scriptsize,inner sep=3pt] at ({\MX-0.5*\MS},\yy) {\nm};
    \foreach \j/\v in {0/\ra,1/\rb,2/\rc,3/\rd}{
      \node[font=\footnotesize] at ({\MX+\j*\MS},\yy) {\v};}
    \node[anchor=west,font=\scriptsize,inner sep=3pt] at ({\MX+3.6*\MS},\yy)
      {\ok\ of 4};
    \node[anchor=west,font=\footnotesize\bfseries,lost,inner sep=3pt]
      at ({\MX+5.3*\MS},\yy) {\pay};}
  \node[anchor=west,font=\scriptsize,black!60,inner sep=3pt]
    at ({\MX+5.15*\MS},{1.20*\RS}) {paid};
  \node[anchor=west,font=\scriptsize,black!55,inner sep=3pt] at (-0.60,{-1.60*\RS})
    {one question, every rung of the ladder};
\end{scope}
\begin{scope}[shift={($(kbot)+(9.15,-2.37)$)}]
  \def\CW{3.35}\def\lo{0.86}\def\hi{0.99}\def\rs{0.44}
  \newcommand{\mx}[1]{{(#1-\lo)/(\hi-\lo)*\CW}}
  \node[anchor=south west,inner sep=0pt] at (-2.10,{3.42*\rs})
    {\textbf{THPT-Ladder}: 632 items, 21 official exams};
  \draw[black!70,line width=.55pt] (-2.06,{2.96*\rs}) circle (0.068);
  \node[anchor=west,font=\scriptsize,black!70,inner sep=3pt] at (-1.98,{2.96*\rs})
    {stmt.\ accuracy};
  \filldraw[black] (0.40,{2.96*\rs}) circle (0.068);
  \node[anchor=west,font=\scriptsize,inner sep=3pt] at (0.48,{2.96*\rs}) {mark earned};
  \draw[{Latex[length=1.2mm]}-,lost,line width=.8pt] (2.30,{2.96*\rs}) -- ++(0.34,0);
  \node[anchor=west,font=\scriptsize,lost,inner sep=3pt] at (2.72,{2.96*\rs}) {shortfall};
  \foreach \i/\nm/\a/\m/\d in {0/{Claude Opus 5}/0.973/0.949/{0.024},
                               1/{GPT-5.5}/0.967/0.948/{0.020},
                               2/{Claude Sonnet 5}/0.935/0.881/{0.054},
                               3/{Qwen3.5-27B}/0.926/0.884/{0.042}}{
    \pgfmathsetmacro\yy{(1.95-\i*0.78)*\rs}
    \node[anchor=east,font=\scriptsize,inner sep=3pt] at (-0.16,\yy) {\nm};
    \draw[{Latex[length=1.15mm]}-,lost,line width=.8pt] (\mx{\m},\yy) -- (\mx{\a},\yy);
    \draw[black!70,fill=white,line width=.55pt] (\mx{\a},\yy) circle (0.068);
    \filldraw[black] (\mx{\m},\yy) circle (0.068);
    \pgfmathsetmacro\xmid{((\a+\m)/2-\lo)/(\hi-\lo)*\CW}
    \node[anchor=south,font=\scriptsize,lost,inner sep=2pt] at (\xmid,\yy) {\d};}
  \draw[black!30,line width=.5pt] (\mx{0.865},{-1.34*\rs}) -- (\mx{0.985},{-1.34*\rs});
  \foreach \t/\lt in {0.88/{0.88},0.92/{0.92},0.96/{0.96}}{
    \node[font=\scriptsize,black!55,below,inner sep=3pt] at (\mx{\t},{-1.34*\rs}) {\lt};}
  \node[font=\scriptsize,black!55,anchor=north,inner sep=1pt] at (\mx{0.765},{-2.20*\rs})
    {per Part~II question (of 1.00)};
\end{scope}
\path let \p1=(kbot) in coordinate (fin) at (0,{\y1-3.72cm});
\draw[black!75,line width=.9pt] (0,0|-fin) -- ({\LW+\RW+0.40},0|-fin);
\end{tikzpicture}}
\caption{One Part~II question from the corpus, as the ministry prints and marks it. The Vietnamese original is on the left, and our English translation is on the right. All four statements are true (\emph{Đ} = \emph{đúng}). This single question demonstrates the convex grading ladder: judging three out of four statements correctly pays 0.50 points, not the 0.75 points that proportional credit would award. The right panel shows how this rule leaves all eight models short by 0.020 to 0.159 points per Part~II question across the corpus. Four models are drawn, showing that their ranking by shortfall differs from their ranking by statement accuracy.}
\label{fig:item}
\end{figure}

The structure is uniform. Decision~764 sets the same three parts for every exam, and Circular 24/2024~\cite{tt24} governs how students sit for them. Part~I consists of four-option multiple choice questions worth 0.25 points each. Part~II groups four true/false statements into one question graded on the convex ladder (Fig.~\ref{fig:item} shows how the ministry prints and marks one). Part~III requires short answers with no options provided. Table~\ref{tab:corpus} details this structure per subject, the exact random baseline it creates, and our coverage. The 2026 exams share this structure because the ministry retained the decision for a second year. Part~II appears in ten of the eleven subjects. Where it appears, its four questions account for 4.00 of the 10.00 total points. This means 40\% of a candidate's mark rests on just four out of the 22 to 28 questions on the exam. The random baseline for an entire exam is therefore not a flat 0.25, nor is it the same across subjects. It is lowest in Mathematics because its large Part~III pays a guesser nothing. The THPT-Ladder corpus holds 632 items from 21 exams across 11 subjects over two years, covering 22 official variants (\emph{mã đề}). Every item carries a direct ministry answer, and the dataset includes 80 figure crops for the 77 items that print a figure.

We count a Part~II question as one single item rather than four, since the ladder pays on the question level. Therefore, our Mathematics exams hold 22 items, even though the ministry's summary reports 34. Informatics prints six Part~II questions but only marks four (Table~\ref{tab:corpus}\textsuperscript{a}) because it includes two common questions and two pairs from elective streams, ensuring its total also sums to 10.00 points.

\begin{table}[t] \caption{Official structure (Decision~764) and corpus coverage.} \label{tab:corpus} \centering \small 
\begin{tabular}{@{}lrrrr rr r@{}}
\toprule
& \multicolumn{4}{c}{structure} & \multicolumn{2}{c}{items} & \\
\cmidrule(lr){2-5}\cmidrule(lr){6-7}
Subject & I & II & III & rnd & 2025 & 2026 & Fig. \\
\midrule
Biology & 18 & 4 & 6 & 2.350 & 28 & 28 & 28 \\
Chemistry & 18 & 4 & 6 & 2.350 & 28 & 28 & 9 \\
Economics \& Law & 24 & 4 & 0 & 2.725 & 28 & 28 & 0 \\
English & 40 & 0 & 0 & 2.500 & 40 & 40 & 0 \\
Geography & 18 & 4 & 6 & 2.350 & 28 & 28 & 8 \\
History & 24 & 4 & 0 & 2.725 & 28 & 28 & 1 \\
Informatics & 24 & 4\textsuperscript{a} & 0 & 2.725 & 30 & 30 & 0 \\
Mathematics & 12 & 4 & 6 & 1.975 & 22 & 22 & 14 \\
Physics & 18 & 4 & 6 & 2.350 & 28 & 28 & 4 \\
Technology (Agri.) & 24 & 4 & 0 & 2.725 & 56 & 28 & 4 \\
Technology (Ind.) & 24 & 4 & 0 & 2.725 & 28 & \textit{n.p.} & 12 \\
\midrule
\textbf{total} & \multicolumn{4}{c}{} & \multicolumn{2}{c}{\textbf{632} items} & \textbf{80} \\
\bottomrule
\end{tabular}
 \par\vspace{1pt}{\footnotesize\upshape\raggedright \emph{I}/\emph{II}/\emph{III}: questions per part, at 0.25, 1.00 and either 0.50 or 0.25 points; 10.00 total in 50 min (Mathematics 90). \emph{rnd}: exact whole-exam random baseline, Eq.~\ref{eq:random} on Part~II and 0 on Part~III. \emph{2025}/\emph{2026}: items held, per year; a count above one exam's total indicates two variants of the same exam, and \emph{n.p.} that the ministry had not published the exam. \emph{Fig.}: figure crops shipped. \textsuperscript{a}6 printed, 4 answered. The chance baseline is therefore neither 25\% nor one number: it runs from 19.75\% of the scale in Mathematics, whose large Part~III pays a guesser nothing, to 27.25\% in the five subjects that ask 24 Part~I items and no Part~III.\par} \end{table}

\section{Evaluation Methodology} \label{sec:metrics}

We score every model exactly as the ministry scores a candidate. A correct four-option item earns 0.25 points, a Part~II question pays the convex ladder value, and a correct Part~III item receives the subject's short-answer value. A model's headline figure in Table~\ref{tab:results} represents the fraction of available points it successfully earned.

Standard accuracy hides what the ladder penalizes, so we measure the penalty directly. Averaged over the $Q$ Part~II questions a model answered, the shortfall is
\begin{equation}
\mathrm{SF} = Q^{-1}\textstyle\sum_{q=1}^{Q}\left(0.25\,a_q - \mathrm{ladder}[a_q]\right) \geq 0.
\label{eq:pcg}
\end{equation}
The shortfall drops to zero only when a model gets every question completely right or completely wrong. It reaches its maximum penalty of 0.25 points on the questions a candidate only half-knows. Because of this, two models that judge the exact same number of statements correctly can earn entirely different marks. The shortfall characterizes the shape and distribution of a model's knowledge, not just its total size.

A mark on one exam cannot be compared directly to the same mark on another. Even with identical structures and marking rules, the two years of a subject were not equally difficult. For example, a score of 6.00 in Economics \& Law beat only 7.99\% of the field in 2025, but it beat 74.22\% of candidates in 2026. On that specific exam, the number of candidates scoring a perfect 10.00 plummeted from 1{,}451 down to 2. Seven of the eleven subjects became harder between the two years. To handle this, we report every model's mark as a percentile rank against the human candidates who sat the exact same exam. Our cohort records successfully reproduce the ministry's published statistics on 310 out of 312 comparisons.

\begin{figure}[t]\centering
\resizebox{0.70\linewidth}{!}{
\begin{tikzpicture}[x=1cm,y=1cm,font=\footnotesize]
\definecolor{lost}{RGB}{160,32,26}
\def\CW{0.60}\def\CH{0.315}
\foreach \j/\p in {0/{(a)},1/{(b)},2/{(c)},3/{(d)}}{
  \node[anchor=south,inner sep=1.5pt] at ({\j*\CW+\CW/2},{0.14}) {\p};}
\foreach \i/\nm/\pa/\pb/\pc/\pd in {%
 0/{Mathematics 2025}/1.000/0.188/0.375/0.438,
 1/{Mathematics 2026}/1.000/0.750/0.375/0.375,
 2/{Technology (Ind.) 2025}/0.875/0.500/0.625/0.250,
 3/{Physics 2026}/0.786/0.542/0.365/0.495,
 4/{Informatics 2026}/0.667/0.583/0.833/0.167,
 5/{Economics \& Law 2025}/0.651/0.625/0.562/0.536,
 6/{Physics 2025}/0.641/0.667/0.589/0.729,
 7/{Technology (Agri.) 2026}/0.615/0.432/0.432/0.646,
 8/{Technology (Agri.) 2025}/0.594/0.495/0.568/0.594,
 9/{Informatics 2025}/0.583/0.500/0.417/0.583,
 10/{Chemistry 2026}/0.578/0.240/0.438/0.620,
 11/{Biology 2025}/0.562/0.750/0.438/0.688,
 12/{Chemistry 2025}/0.552/0.521/0.557/0.495,
 13/{Technology (Ind.) 2026}/0.500/0.750/0.500/0.250,
 14/{History 2025}/0.484/0.688/0.406/0.422,
 15/{Geography 2025}/0.448/0.521/0.396/0.510,
 16/{History 2026}/0.448/0.484/0.589/0.479,
 17/{Biology 2026}/0.375/0.625/0.750/0.250,
 18/{Economics \& Law 2026}/0.370/0.464/0.354/0.312,
 19/{Geography 2026}/0.286/0.740/0.333/0.141}{
  \pgfmathsetmacro\ry{-\i*\CH}
  \node[anchor=east,inner sep=3pt] at (-0.06,{\ry-\CH/2}) {\nm};
  \foreach \j/\v in {0/\pa,1/\pb,2/\pc,3/\pd}{
    \pgfmathsetmacro\sh{\v*88}
    \pgfmathsetmacro\dark{ifthenelse(\v>0.66,1,0)}
    \fill[lost!\sh!white] ({\j*\CW},{\ry-\CH}) rectangle ++(\CW,\CH);
    \draw[white,line width=.5pt] ({\j*\CW},{\ry-\CH}) rectangle ++(\CW,\CH);
    \ifnum\dark=1
      \node[font=\scriptsize,white] at ({\j*\CW+\CW/2},{\ry-\CH/2}) {\pgfmathprintnumber[fixed,precision=2,fixed zerofill]{\v}};
    \else
      \node[font=\scriptsize] at ({\j*\CW+\CW/2},{\ry-\CH/2}) {\pgfmathprintnumber[fixed,precision=2,fixed zerofill]{\v}};
    \fi}}
\node[anchor=north,inner sep=4pt] at ({2*\CW},{-20*\CH}) {statement position};
\draw[lost,line width=.6pt] (-0.02,{-0*\CH+0.02}) rectangle ({4*\CW+0.02},{-2*\CH-0.02});
\draw[lost,line width=.6pt] (-0.02,{-19*\CH+0.02}) rectangle ({4*\CW+0.02},{-20*\CH-0.02});

\end{tikzpicture}}
\caption{How often each Part~II statement is true across the twenty subject-years where we hold the keys. In Mathematics, position (a) is always true in all 96 questions every year. However, in Geography 2026, position (a) is true in only 29\% of questions and position (d) in 14\%. This shows that the answer keys are not statement-balanced and that the imbalance changes direction depending on the subject. This imbalance allows a fixed answer string to beat random chance without reading a single question.}
\label{fig:keys}
\end{figure}

\subsection{Answer-Only Baseline} \label{sec:floor}

Eq.~\ref{eq:random} assumes a candidate decides each of a question's four statements independently. However, a respondent who simply submits the same four-value string to every question exploits the imbalance between true and false answers in the official key. We tested whether the published keys carry exploitable spurious features by scoring strings chosen without looking at the exam text. When we select the string \texttt{DDSS} based on the other 19 subject-years and apply it unseen, it wins on all 20 exams. It earns between 11.07\% and 24.25\% of a ten-point exam, beating independent guessing on 18 of them. It represents the most frequent of the sixteen possible patterns, appearing 14.5\% of the time compared to the 6.25\% a perfectly balanced key would produce. Furthermore, the imbalance runs in opposite directions by subject\ (Fig.~\ref{fig:keys}). A Part~II score at or below this level demonstrates no competence at all.

\section{Experiments} \label{sec:experiments}

We evaluate eight models in Vietnamese, presenting one item at a time with its text and figures and no
worked solution, under prompts identical across subjects and years. Three are open-weight checkpoints
served locally at their publishers' recommended sampling settings, capped at 16{,}384 generated tokens by
the memory available; five are closed models from two vendors, and all but one of those rejects any sampling
setting but its default, so the arms are not matched on decoding. Six open-weight responses reach that cap or
cannot be read and score nothing, so the Part~II comparisons are also reported over the 82 questions free of
them. Every one of the 632 items is scored against the ministry's key by the marking rule
of Section~\ref{sec:metrics}.

\subsection{A mark locates a model in the cohort, unevenly}

Because the ministry publishes the score distribution for every exam, we can read a mark as a true rank among the human candidates the examination was written for, rather than against an arbitrary scale. On the eighteen subject-years where we hold a complete exam, Qwen3.5-27B ranks above 99.95\% of the 1{,}126{,}172 candidates who sat Mathematics 2025, but only above 76.97\% of those who sat History 2025. InternVL3.5-8B fluctuates even more, running from the 17.8th percentile up to the 97.9th\ (Fig.~\ref{fig:pct}). A single average score completely conceals a range this wide. Meanwhile, Claude Opus 5 and GPT-5.5 score 9.67 and 9.64 out of 10, respectively, with each taking full marks on eight of the eighteen exams. At this level of performance, the aggregate mark is nearly exhausted as a useful measurement instrument, but the partial-credit shortfall still differentiates them.

\subsection{The exams a cohort finds hard are not the exams a model finds hard}

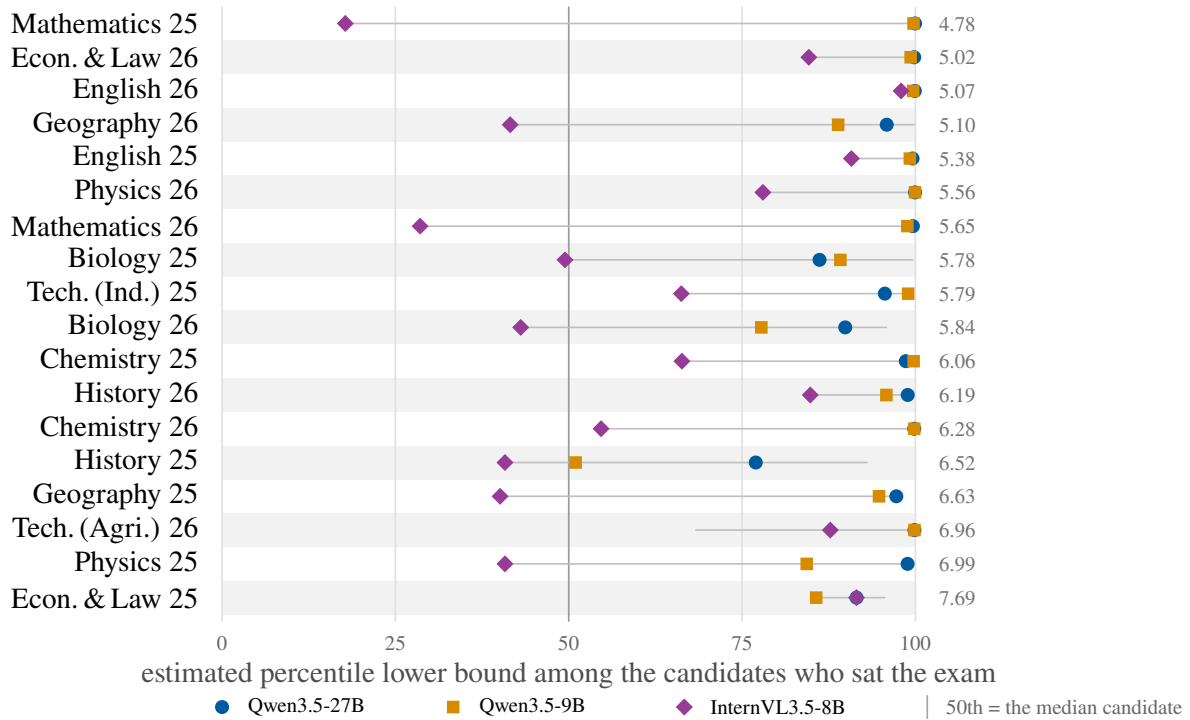
\begin{figure}[t]\centering
\resizebox{\linewidth}{!}{
\begin{tikzpicture}[x=1cm,y=1cm]
\footnotesize
\definecolor{cQwen3527B}{RGB}{0,90,160}
\definecolor{cQwen359B}{RGB}{215,140,0}
\definecolor{cInternVL358B}{RGB}{150,60,150}
\fill[black!5] (0,4.560) rectangle (5.850,4.845);
\fill[black!5] (0,3.990) rectangle (5.850,4.275);
\fill[black!5] (0,3.420) rectangle (5.850,3.705);
\fill[black!5] (0,2.850) rectangle (5.850,3.135);
\fill[black!5] (0,2.280) rectangle (5.850,2.565);
\fill[black!5] (0,1.710) rectangle (5.850,1.995);
\fill[black!5] (0,1.140) rectangle (5.850,1.425);
\fill[black!5] (0,0.570) rectangle (5.850,0.855);
\fill[black!5] (0,0.000) rectangle (5.850,0.285);
\draw[black!12] (0.000,-0.060) -- (0.000,5.130);
\node[font=\tiny,anchor=north,text=black!60] at (0.000,-0.080) {0};
\draw[black!12] (1.462,-0.060) -- (1.462,5.130);
\node[font=\tiny,anchor=north,text=black!60] at (1.462,-0.080) {25};
\draw[black!40] (2.925,-0.060) -- (2.925,5.130);
\node[font=\tiny,anchor=north,text=black!60] at (2.925,-0.080) {50};
\draw[black!12] (4.387,-0.060) -- (4.387,5.130);
\node[font=\tiny,anchor=north,text=black!60] at (4.387,-0.080) {75};
\draw[black!12] (5.850,-0.060) -- (5.850,5.130);
\node[font=\tiny,anchor=north,text=black!60] at (5.850,-0.080) {100};
\node[font=\scriptsize,anchor=north,text=black!70] at (2.925,-0.300) {estimated percentile lower bound among the candidates who sat the exam};
\node[font=\scriptsize,anchor=east] at (-0.100,4.987) {Mathematics 25};
\node[font=\tiny,anchor=west,text=black!55] at (5.950,4.987) {4.78};
\draw[black!25,line width=0.4pt] (1.040,4.987) -- (5.847,4.987);
\filldraw[cQwen3527B] (5.847,4.987) circle (0.052);
\filldraw[cQwen359B] (5.788,4.941) rectangle (5.880,5.034);
\filldraw[cInternVL358B] (1.040,5.050) -- (1.102,4.987) -- (1.040,4.925) -- (0.978,4.987) -- cycle;
\node[font=\scriptsize,anchor=east] at (-0.100,4.702) {Econ.\,\&\,Law 26};
\node[font=\tiny,anchor=west,text=black!55] at (5.950,4.702) {5.02};
\draw[black!25,line width=0.4pt] (4.951,4.702) -- (5.849,4.702);
\filldraw[cQwen3527B] (5.840,4.702) circle (0.052);
\filldraw[cQwen359B] (5.764,4.656) rectangle (5.856,4.748);
\filldraw[cInternVL358B] (4.951,4.764) -- (5.013,4.702) -- (4.951,4.640) -- (4.889,4.702) -- cycle;
\node[font=\scriptsize,anchor=east] at (-0.100,4.418) {English 26};
\node[font=\tiny,anchor=west,text=black!55] at (5.950,4.418) {5.07};
\draw[black!25,line width=0.4pt] (5.674,4.418) -- (5.845,4.418);
\filldraw[cQwen3527B] (5.845,4.418) circle (0.052);
\filldraw[cQwen359B] (5.785,4.372) rectangle (5.877,4.464);
\filldraw[cInternVL358B] (5.729,4.480) -- (5.791,4.418) -- (5.729,4.356) -- (5.667,4.418) -- cycle;
\node[font=\scriptsize,anchor=east] at (-0.100,4.133) {Geography 26};
\node[font=\tiny,anchor=west,text=black!55] at (5.950,4.133) {5.10};
\draw[black!25,line width=0.4pt] (2.431,4.133) -- (5.847,4.133);
\filldraw[cQwen3527B] (5.608,4.133) circle (0.052);
\filldraw[cQwen359B] (5.152,4.087) rectangle (5.244,4.179);
\filldraw[cInternVL358B] (2.431,4.195) -- (2.493,4.133) -- (2.431,4.071) -- (2.369,4.133) -- cycle;
\node[font=\scriptsize,anchor=east] at (-0.100,3.848) {English 25};
\node[font=\tiny,anchor=west,text=black!55] at (5.950,3.848) {5.38};
\draw[black!25,line width=0.4pt] (5.310,3.848) -- (5.848,3.848);
\filldraw[cQwen3527B] (5.826,3.848) circle (0.052);
\filldraw[cQwen359B] (5.757,3.802) rectangle (5.849,3.893);
\filldraw[cInternVL358B] (5.310,3.909) -- (5.372,3.848) -- (5.310,3.786) -- (5.248,3.848) -- cycle;
\node[font=\scriptsize,anchor=east] at (-0.100,3.562) {Physics 26};
\node[font=\tiny,anchor=west,text=black!55] at (5.950,3.562) {5.56};
\draw[black!25,line width=0.4pt] (4.563,3.562) -- (5.847,3.562);
\filldraw[cQwen3527B] (5.847,3.562) circle (0.052);
\filldraw[cQwen359B] (5.801,3.517) rectangle (5.893,3.608);
\filldraw[cInternVL358B] (4.563,3.624) -- (4.625,3.562) -- (4.563,3.501) -- (4.501,3.562) -- cycle;
\node[font=\scriptsize,anchor=east] at (-0.100,3.277) {Mathematics 26};
\node[font=\tiny,anchor=west,text=black!55] at (5.950,3.277) {5.65};
\draw[black!25,line width=0.4pt] (1.671,3.277) -- (5.829,3.277);
\filldraw[cQwen3527B] (5.829,3.277) circle (0.052);
\filldraw[cQwen359B] (5.736,3.232) rectangle (5.828,3.323);
\filldraw[cInternVL358B] (1.671,3.339) -- (1.733,3.277) -- (1.671,3.216) -- (1.609,3.277) -- cycle;
\node[font=\scriptsize,anchor=east] at (-0.100,2.993) {Biology 25};
\node[font=\tiny,anchor=west,text=black!55] at (5.950,2.993) {5.78};
\draw[black!25,line width=0.4pt] (2.893,2.993) -- (5.833,2.993);
\filldraw[cQwen3527B] (5.041,2.993) circle (0.052);
\filldraw[cQwen359B] (5.171,2.947) rectangle (5.263,3.038);
\filldraw[cInternVL358B] (2.893,3.054) -- (2.955,2.993) -- (2.893,2.931) -- (2.831,2.993) -- cycle;
\node[font=\scriptsize,anchor=east] at (-0.100,2.708) {Tech.\,(Ind.) 25};
\node[font=\tiny,anchor=west,text=black!55] at (5.950,2.708) {5.79};
\draw[black!25,line width=0.4pt] (3.875,2.708) -- (5.827,2.708);
\filldraw[cQwen3527B] (5.592,2.708) circle (0.052);
\filldraw[cQwen359B] (5.743,2.662) rectangle (5.835,2.753);
\filldraw[cInternVL358B] (3.875,2.769) -- (3.937,2.708) -- (3.875,2.646) -- (3.813,2.708) -- cycle;
\node[font=\scriptsize,anchor=east] at (-0.100,2.423) {Biology 26};
\node[font=\tiny,anchor=west,text=black!55] at (5.950,2.423) {5.84};
\draw[black!25,line width=0.4pt] (2.520,2.423) -- (5.611,2.423);
\filldraw[cQwen3527B] (5.258,2.423) circle (0.052);
\filldraw[cQwen359B] (4.504,2.377) rectangle (4.596,2.469);
\filldraw[cInternVL358B] (2.520,2.485) -- (2.582,2.423) -- (2.520,2.361) -- (2.458,2.423) -- cycle;
\node[font=\scriptsize,anchor=east] at (-0.100,2.138) {Chemistry 25};
\node[font=\tiny,anchor=west,text=black!55] at (5.950,2.138) {6.06};
\draw[black!25,line width=0.4pt] (3.880,2.138) -- (5.835,2.138);
\filldraw[cQwen3527B] (5.769,2.138) circle (0.052);
\filldraw[cQwen359B] (5.789,2.092) rectangle (5.881,2.183);
\filldraw[cInternVL358B] (3.880,2.200) -- (3.942,2.138) -- (3.880,2.076) -- (3.818,2.138) -- cycle;
\node[font=\scriptsize,anchor=east] at (-0.100,1.853) {History 26};
\node[font=\tiny,anchor=west,text=black!55] at (5.950,1.853) {6.19};
\draw[black!25,line width=0.4pt] (4.963,1.853) -- (5.785,1.853);
\filldraw[cQwen3527B] (5.785,1.853) circle (0.052);
\filldraw[cQwen359B] (5.559,1.806) rectangle (5.651,1.899);
\filldraw[cInternVL358B] (4.963,1.915) -- (5.025,1.853) -- (4.963,1.790) -- (4.901,1.853) -- cycle;
\node[font=\scriptsize,anchor=east] at (-0.100,1.568) {Chemistry 26};
\node[font=\tiny,anchor=west,text=black!55] at (5.950,1.568) {6.28};
\draw[black!25,line width=0.4pt] (3.198,1.568) -- (5.840,1.568);
\filldraw[cQwen3527B] (5.840,1.568) circle (0.052);
\filldraw[cQwen359B] (5.794,1.522) rectangle (5.886,1.614);
\filldraw[cInternVL358B] (3.198,1.630) -- (3.260,1.568) -- (3.198,1.506) -- (3.136,1.568) -- cycle;
\node[font=\scriptsize,anchor=east] at (-0.100,1.283) {History 25};
\node[font=\tiny,anchor=west,text=black!55] at (5.950,1.283) {6.52};
\draw[black!25,line width=0.4pt] (2.387,1.283) -- (5.448,1.283);
\filldraw[cQwen3527B] (4.503,1.283) circle (0.052);
\filldraw[cQwen359B] (2.937,1.237) rectangle (3.029,1.329);
\filldraw[cInternVL358B] (2.387,1.345) -- (2.449,1.283) -- (2.387,1.221) -- (2.325,1.283) -- cycle;
\node[font=\scriptsize,anchor=east] at (-0.100,0.998) {Geography 25};
\node[font=\tiny,anchor=west,text=black!55] at (5.950,0.998) {6.63};
\draw[black!25,line width=0.4pt] (2.346,0.998) -- (5.689,0.998);
\filldraw[cQwen3527B] (5.689,0.998) circle (0.052);
\filldraw[cQwen359B] (5.497,0.952) rectangle (5.589,1.044);
\filldraw[cInternVL358B] (2.346,1.060) -- (2.408,0.998) -- (2.346,0.936) -- (2.284,0.998) -- cycle;
\node[font=\scriptsize,anchor=east] at (-0.100,0.713) {Tech.\,(Agri.) 26};
\node[font=\tiny,anchor=west,text=black!55] at (5.950,0.713) {6.96};
\draw[black!25,line width=0.4pt] (3.993,0.713) -- (5.842,0.713);
\filldraw[cQwen3527B] (5.842,0.713) circle (0.052);
\filldraw[cQwen359B] (5.796,0.667) rectangle (5.888,0.759);
\filldraw[cInternVL358B] (5.132,0.775) -- (5.194,0.713) -- (5.132,0.651) -- (5.070,0.713) -- cycle;
\node[font=\scriptsize,anchor=east] at (-0.100,0.428) {Physics 25};
\node[font=\tiny,anchor=west,text=black!55] at (5.950,0.428) {6.99};
\draw[black!25,line width=0.4pt] (2.387,0.428) -- (5.784,0.428);
\filldraw[cQwen3527B] (5.784,0.428) circle (0.052);
\filldraw[cQwen359B] (4.887,0.382) rectangle (4.979,0.474);
\filldraw[cInternVL358B] (2.387,0.490) -- (2.449,0.428) -- (2.387,0.366) -- (2.325,0.428) -- cycle;
\node[font=\scriptsize,anchor=east] at (-0.100,0.143) {Econ.\,\&\,Law 25};
\node[font=\tiny,anchor=west,text=black!55] at (5.950,0.143) {7.69};
\draw[black!25,line width=0.4pt] (5.013,0.143) -- (5.596,0.143);
\filldraw[cQwen3527B] (5.353,0.143) circle (0.052);
\filldraw[cQwen359B] (4.967,0.097) rectangle (5.059,0.189);
\filldraw[cInternVL358B] (5.353,0.205) -- (5.415,0.143) -- (5.353,0.081) -- (5.291,0.143) -- cycle;
\filldraw[cQwen3527B] (0.000,-0.780) circle (0.052);
\node[font=\tiny,anchor=west] at (0.120,-0.780) {Qwen3.5-27B};
\filldraw[cQwen359B] (1.904,-0.826) rectangle (1.996,-0.734);
\node[font=\tiny,anchor=west] at (2.070,-0.780) {Qwen3.5-9B};
\filldraw[cInternVL358B] (3.900,-0.718) -- (3.962,-0.780) -- (3.900,-0.842) -- (3.838,-0.780) -- cycle;
\node[font=\tiny,anchor=west] at (4.020,-0.780) {InternVL3.5-8B};
\draw[black!35] (5.950,-0.880) -- (5.950,-0.680);
\node[font=\tiny,anchor=west,text=black!60] at (6.010,-0.780) {50th = the median candidate};
\end{tikzpicture}}
\caption{Each open-weight model's rank among the candidates who sat the same exam, for the
eighteen subject-years where the corpus holds a whole 10.00-mark exam; the closed models'
standings are given in the text. Rows run from the exam the
cohort found hardest to the one it found easiest, with the cohort's own mean mark at the right.
Percentiles are lower bounds: a mark on the ministry's 0.25 grid ties with every candidate who
earned it. If difficulty transferred from candidates to models the markers would drift
rightwards down the figure; they do not, which indicates that a model's standing is set by
something other than what the exam cost a human.}
\label{fig:pct}
\end{figure}

\begin{figure}[t]\centering
\resizebox{0.82\linewidth}{!}{
\begin{tikzpicture}[x=1cm,y=1cm]
\footnotesize
\definecolor{dQwen3527B}{RGB}{0,90,160}
\definecolor{dQwen359B}{RGB}{215,140,0}
\definecolor{dInternVL358B}{RGB}{150,60,150}
\draw[black!25] (0,0) rectangle (4.050,3.000);
\draw[black!8] (0.054,0) -- (0.054,3.000);
\draw[black!8] (0,0.040) -- (4.050,0.040);
\draw[black!8] (0.598,0) -- (0.598,3.000);
\draw[black!8] (0,0.443) -- (4.050,0.443);
\draw[black!8] (1.142,0) -- (1.142,3.000);
\draw[black!8] (0,0.846) -- (4.050,0.846);
\draw[black!8] (1.685,0) -- (1.685,3.000);
\draw[black!8] (0,1.248) -- (4.050,1.248);
\draw[black!8] (2.229,0) -- (2.229,3.000);
\draw[black!8] (0,1.651) -- (4.050,1.651);
\draw[black!8] (2.772,0) -- (2.772,3.000);
\draw[black!8] (0,2.054) -- (4.050,2.054);
\draw[black!8] (3.316,0) -- (3.316,3.000);
\draw[black!8] (0,2.456) -- (4.050,2.456);
\draw[black!8] (3.860,0) -- (3.860,3.000);
\draw[black!8] (0,2.859) -- (4.050,2.859);
\node[font=\tiny,anchor=north,text=black!60] at (0.598,-0.050) {4};
\node[font=\tiny,anchor=east,text=black!60] at (-0.050,0.443) {4};
\node[font=\tiny,anchor=north,text=black!60] at (1.685,-0.050) {6};
\node[font=\tiny,anchor=east,text=black!60] at (-0.050,1.248) {6};
\node[font=\tiny,anchor=north,text=black!60] at (2.772,-0.050) {8};
\node[font=\tiny,anchor=east,text=black!60] at (-0.050,2.054) {8};
\node[font=\tiny,anchor=north,text=black!60] at (3.860,-0.050) {10};
\node[font=\tiny,anchor=east,text=black!60] at (-0.050,2.859) {10};
\draw[black!45,dashed,line width=0.6pt] (0.000,0.000) -- (4.050,3.000);
\node[font=\tiny,text=black!50,rotate=45,anchor=south] at (3.099,2.295) {equal marks};
\draw[dQwen3527B!55,line width=0.45pt,dash pattern=on 2pt off 1.5pt] (0,2.579) -- (4.050,2.579);
\draw[dQwen359B!55,line width=0.45pt,dash pattern=on 2pt off 1.5pt] (0,2.450) -- (4.050,2.450);
\draw[dInternVL358B!55,line width=0.45pt,dash pattern=on 2pt off 1.5pt] (0,1.515) -- (4.050,1.515);
\filldraw[dQwen3527B] (1.572,2.356) circle (0.048);
\filldraw[dQwen3527B] (2.208,2.859) circle (0.048);
\filldraw[dQwen3527B] (2.027,2.758) circle (0.048);
\filldraw[dQwen3527B] (1.196,2.054) circle (0.048);
\filldraw[dQwen3527B] (2.605,2.557) circle (0.048);
\filldraw[dQwen3527B] (1.154,2.356) circle (0.048);
\filldraw[dQwen3527B] (1.721,2.758) circle (0.048);
\filldraw[dQwen3527B] (1.840,2.859) circle (0.048);
\filldraw[dQwen3527B] (1.968,2.054) circle (0.048);
\filldraw[dQwen3527B] (1.789,2.758) circle (0.048);
\filldraw[dQwen3527B] (1.345,2.658) circle (0.048);
\filldraw[dQwen3527B] (1.182,2.859) circle (0.048);
\filldraw[dQwen3527B] (1.565,1.953) circle (0.048);
\filldraw[dQwen3527B] (1.597,2.154) circle (0.048);
\filldraw[dQwen3527B] (1.024,2.859) circle (0.048);
\filldraw[dQwen3527B] (1.496,2.859) circle (0.048);
\filldraw[dQwen3527B] (2.221,2.859) circle (0.048);
\filldraw[dQwen3527B] (1.445,2.859) circle (0.048);
\filldraw[dQwen359B] (1.530,2.616) rectangle (1.614,2.700);
\filldraw[dQwen359B] (2.166,2.817) rectangle (2.250,2.901);
\filldraw[dQwen359B] (1.985,2.616) rectangle (2.069,2.700);
\filldraw[dQwen359B] (1.154,1.710) rectangle (1.238,1.794);
\filldraw[dQwen359B] (2.563,2.414) rectangle (2.647,2.498);
\filldraw[dQwen359B] (1.112,2.112) rectangle (1.196,2.196);
\filldraw[dQwen359B] (1.679,2.817) rectangle (1.763,2.901);
\filldraw[dQwen359B] (1.798,2.817) rectangle (1.882,2.901);
\filldraw[dQwen359B] (1.926,1.508) rectangle (2.010,1.592);
\filldraw[dQwen359B] (1.747,2.515) rectangle (1.831,2.599);
\filldraw[dQwen359B] (1.303,2.515) rectangle (1.387,2.599);
\filldraw[dQwen359B] (1.140,2.716) rectangle (1.224,2.800);
\filldraw[dQwen359B] (1.523,2.052) rectangle (1.607,2.136);
\filldraw[dQwen359B] (1.555,1.710) rectangle (1.639,1.794);
\filldraw[dQwen359B] (0.982,2.616) rectangle (1.066,2.700);
\filldraw[dQwen359B] (1.454,2.616) rectangle (1.538,2.700);
\filldraw[dQwen359B] (2.179,2.354) rectangle (2.263,2.438);
\filldraw[dQwen359B] (1.403,2.817) rectangle (1.487,2.901);
\filldraw[dInternVL358B] (1.572,1.506) -- (1.628,1.410) -- (1.516,1.410) -- cycle;
\filldraw[dInternVL358B] (2.208,2.311) -- (2.264,2.215) -- (2.152,2.215) -- cycle;
\filldraw[dInternVL358B] (2.027,1.405) -- (2.083,1.309) -- (1.971,1.309) -- cycle;
\filldraw[dInternVL358B] (1.196,0.801) -- (1.252,0.705) -- (1.140,0.705) -- cycle;
\filldraw[dInternVL358B] (2.605,2.613) -- (2.661,2.517) -- (2.549,2.517) -- cycle;
\filldraw[dInternVL358B] (1.154,1.546) -- (1.210,1.450) -- (1.098,1.450) -- cycle;
\filldraw[dInternVL358B] (1.721,1.707) -- (1.777,1.611) -- (1.665,1.611) -- cycle;
\filldraw[dInternVL358B] (1.840,1.606) -- (1.896,1.510) -- (1.784,1.510) -- cycle;
\filldraw[dInternVL358B] (1.968,1.405) -- (2.024,1.309) -- (1.912,1.309) -- cycle;
\filldraw[dInternVL358B] (1.789,2.251) -- (1.845,2.155) -- (1.733,2.155) -- cycle;
\filldraw[dInternVL358B] (1.345,1.908) -- (1.401,1.812) -- (1.289,1.812) -- cycle;
\filldraw[dInternVL358B] (1.182,2.512) -- (1.238,2.416) -- (1.126,2.416) -- cycle;
\filldraw[dInternVL358B] (1.565,1.204) -- (1.621,1.108) -- (1.509,1.108) -- cycle;
\filldraw[dInternVL358B] (1.597,1.143) -- (1.653,1.047) -- (1.541,1.047) -- cycle;
\filldraw[dInternVL358B] (1.024,0.197) -- (1.080,0.101) -- (0.968,0.101) -- cycle;
\filldraw[dInternVL358B] (1.496,0.741) -- (1.552,0.645) -- (1.440,0.645) -- cycle;
\filldraw[dInternVL358B] (2.221,1.606) -- (2.277,1.510) -- (2.165,1.510) -- cycle;
\filldraw[dInternVL358B] (1.445,1.808) -- (1.501,1.712) -- (1.389,1.712) -- cycle;
\draw[black!40,line width=0.5pt,dotted] (1.024,0) -- (1.024,-0.275);
\node[font=\tiny,text=black!70,anchor=north,inner sep=1pt] at (1.024,-0.265) {Math\,25};
\draw[black!40,line width=0.5pt,dotted] (2.605,0) -- (2.605,-0.275);
\node[font=\tiny,text=black!70,anchor=north,inner sep=1pt] at (2.605,-0.265) {E\&L\,25};
\node[font=\scriptsize,anchor=north,text=black!75] at (2.025,-0.470) {candidates' average mark};
\node[font=\scriptsize,rotate=90,anchor=south,text=black!75] at (-0.340,1.500) {model's mark};
\filldraw[dQwen3527B] (0.060,-0.860) circle (0.048);
\node[font=\tiny,anchor=west,inner sep=2pt] at (0.14,-0.860) {Qwen3.5-27B\ \ $\rho=0.04$};
\filldraw[dQwen359B] (0.018,-1.122) rectangle (0.102,-1.038);
\node[font=\tiny,anchor=west,inner sep=2pt] at (0.14,-1.080) {Qwen3.5-9B\ \ $\rho=0.02$};
\filldraw[dInternVL358B] (0.060,-1.244) -- (0.116,-1.340) -- (0.004,-1.340) -- cycle;
\node[font=\tiny,anchor=west,inner sep=2pt] at (0.14,-1.300) {InternVL3.5-8B\ \ $\rho=0.35$};
\end{tikzpicture}}
\caption{The same eighteen exams, with the cohort's mean mark against each open-weight model's
on one 0--10 scale; the closed models' correlations are given in the text. The diagonal marks
equal marks; each dashed horizontal is one model's own mean. A
cloud running parallel to the diagonal would mean difficulty transfers from candidates to models,
and a cloud along a horizontal that it does not. Every cloud here is flat: a model's mark is therefore
near-independent of what the exam cost its candidates. No cohort averaged above 7.7, which is why the upper right
is empty. Labelled: the cohort's hardest exam and its easiest.}
\label{fig:diff}
\end{figure}
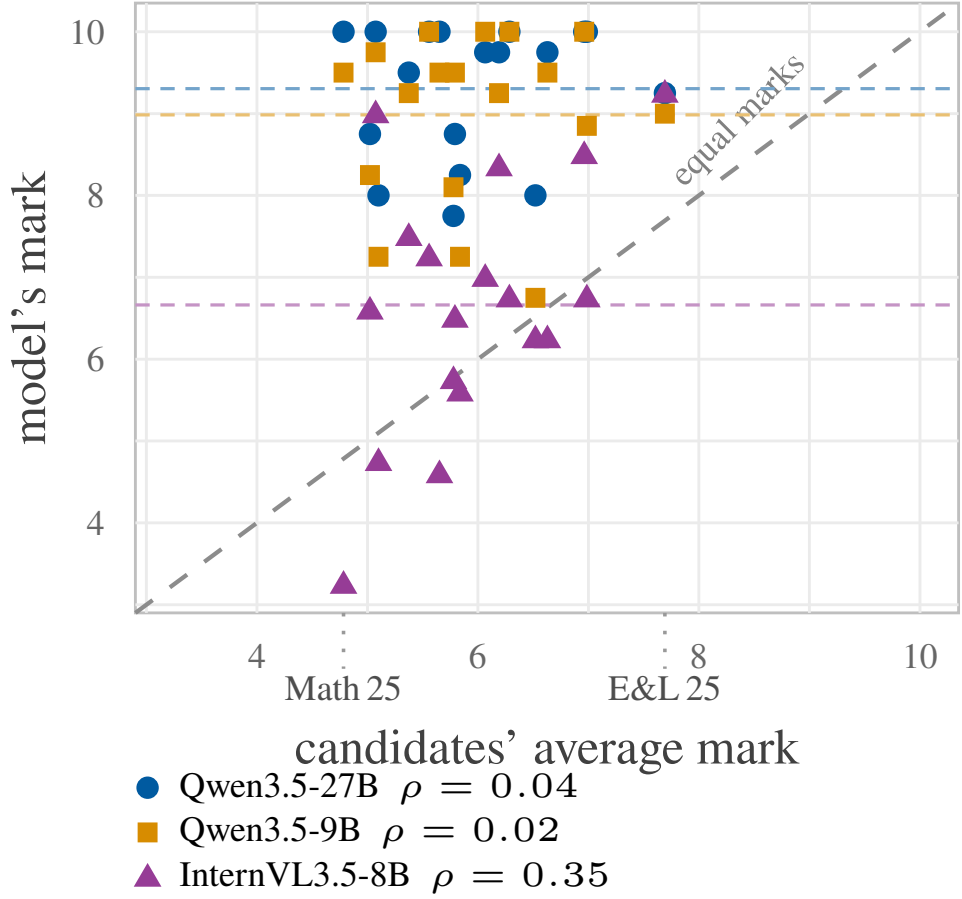

When we rank the eighteen exams by the mean mark the human cohort earned versus the mark each model earned\ (Fig.~\ref{fig:diff}), the two orderings show no detectable agreement. The Spearman correlation ($\rho$) is 0.04 for Qwen3.5-27B, 0.02 for Qwen3.5-9B, and 0.35 for InternVL3.5-8B. None of these correlations are statistically significant, and the first two are indistinguishable from random noise. While previous studies have examined whether a model's difficulty aligns with humans at the individual item level~\cite{psychalign}, our unit of measurement is the full exam, matching how the state defines and publishes the marks. This lack of correlation is not a statistical artifact of our sample size. The exact same eighteen exams detect strong agreement where it genuinely exists: Qwen3.5-27B and GPT-5.5 rank the exams almost identically ($\rho = 0.86$, $p < 0.001$), despite crossing both vendor and licensing boundaries. For human candidates, Mathematics 2025 proved to be the hardest exam in the corpus, averaging 4.78 out of 10, yet Qwen3.5-27B scored a perfect 10.00 on it. Conversely, History 2025 was much easier for candidates (averaging 6.52), but it is where Qwen3.5-27B stands lowest within its cohort\ (Fig.~\ref{fig:pct}). The five closed models fare no better at mimicking human difficulty, with every correlation falling between $-0.14$ and $0.22$ and none reaching significance. Since difficulty for a candidate and difficulty for a model act as nearly independent quantities on this examination, a model's raw score on a subject does not tell us whether that subject is objectively hard.

\begin{table}[t] \caption{Eight models on the 632 scored items: three open-weight, then five closed.} \label{tab:results} \centering \small 
\begin{tabular}{@{}lrrrrrr@{}}
\toprule
Model & \%\,avail. & I & II\,stmt & II\,pts/q & shf/q & III \\
\midrule
Qwen3.5-27B & 93.8 & 0.973 & 0.926 & 0.884 & 0.042 & 0.917 \\
Qwen3.5-9B & 90.7 & 0.965 & 0.896 & 0.830 & 0.066 & 0.850 \\
InternVL3.5-8B & 69.3 & 0.838 & 0.756 & 0.597 & 0.159 & 0.183 \\
Claude Opus 5 & 97.0 & 0.988 & 0.973 & 0.949 & 0.024 & 0.950 \\
GPT-5.5 & 97.0 & 0.988 & 0.967 & 0.948 & 0.020 & 0.950 \\
Claude Sonnet 5 & 93.6 & 0.977 & 0.935 & 0.881 & 0.054 & 0.917 \\
Claude Opus 4.8 & 92.9 & 0.975 & 0.920 & 0.867 & 0.052 & 0.917 \\
Claude Haiku 4.5 & 82.9 & 0.934 & 0.842 & 0.735 & 0.108 & 0.567 \\
\bottomrule
\end{tabular}
 \par\vspace{1pt}{\footnotesize\upshape\raggedright
\emph{\%\,avail.}: the marks a model earned as a percentage of the 220.00 the 22 variants are
worth; Informatics is the mean over its two elective streams. \emph{I}, \emph{III}: the fraction of four-option and short-answer items
answered correctly. \emph{II\,stmt}: the fraction of individual true/false statements judged
correctly. \emph{II\,pts/q}: what the ladder paid, per Part~II question, out of
1.00. \emph{shf/q}: the shortfall $\mathrm{SF}$ of Eq.~\ref{eq:pcg}, the credit the ladder withheld
(lower is better). The open-weight models are served locally at their publishers' recommended
sampling settings. The closed models are Anthropic's Claude Opus~5, Opus~4.8, Sonnet~5 and
Haiku~4.5 and OpenAI's GPT-5.5, reached through Amazon Bedrock; all but Haiku~4.5 reject any
sampling setting but their default. The shortfall is positive for all eight, which shows each
earns fewer marks than its statement accuracy implies, and reveals that the shortfall is not
ordered by that accuracy.\par} \end{table}

\subsection{The shortfall follows the shape of a model's errors}

Every model loses marks to the convex ladder (Table~\ref{tab:results}). If credit were awarded one statement at a time, Qwen3.5-27B would earn 0.926 points per question, but the ladder only pays it 0.884. Across all eight models, this shortfall runs from 0.020 to 0.159 points per question.

Crucially, statement accuracy does not predict this penalty. When we rank models by accuracy, the shortfall does not decrease monotonically. GPT-5.5 knows 96.7\% of statements compared to Claude Opus 5's 97.3\%, yet it faces a smaller penalty (0.020 versus 0.024). Claude Sonnet 5 knows 93.5\% of statements compared to Qwen3.5-27B's 92.6\%, but it suffers a higher penalty (0.054 versus 0.042). If we hold a model's overall accuracy constant and only vary how its errors cluster across questions, the resulting mark becomes a wide interval rather than a single number. For example, at Sonnet 5's accuracy level, the ladder can pay anywhere between 0.869 and 0.932 points per question. Because of this, accuracy simply cannot recover the true mark.

One specific comparison perfectly isolates the ladder's effect. On four occasions, a model and the blind guessing string from Section~\ref{sec:floor} judged the exact same number of statements correctly on the same exam but received entirely different payouts. The marks depend on how the correct statements group together, not on the total count. On the Biology 2026 exam, Qwen3.5-27B and the fixed string each got 11 out of 16 statements right. However, the blind string earned 2.35 points while the model earned only 1.75. The model was uniformly close on every question (getting 3, 2, 3, and 3 statements correct), whereas the blind string swept two questions completely (getting 2, 4, 1, and 4 statements correct). The convex ladder explicitly pays for the second shape.

\section{Implications for Assessment Practice}

Three constraints follow for anyone automating marking on this examination. First, a marking
engine has to implement the ladder of Decision~764 rather than proportional credit: the two
disagree by 0.020 to 0.159 points per Part~II question, and on History 2025 that gap separates
the 90th percentile from the 77th among 481{,}293 candidates.

Second, an accuracy figure does not bound the mark a system would be awarded, because at a fixed
statement accuracy the ladder pays between 0.869 and 0.932 points per question; procurement
evidence should therefore be a mark under the published rules, read against the answer-only
floor of Section~\ref{sec:floor}. Third, model scores are not candidate difficulty data: over
the eighteen exams no correlation between cohort mean marks and model marks reached
significance ($\rho$ from $-0.14$ to $0.35$), so an item bank calibrated on model performance
would not inherit the difficulty ordering its own students experience.

\section{Limitations} \label{sec:limits}

Percentiles are coarser than they initially appear. A single step on the ministry's 0.25 grading grid moves a candidate several percentile places near the median. Therefore, this benchmark cannot separate two models that land within a quarter point of each other. The grid is set by the state and cannot be refined.

Because the 2025 exams have been public for over a year, models may have encountered them during training. Data contamination would normally reveal itself if a model suddenly lost ground on the unseen 2026 exams, assuming we hold the human cohort's change in difficulty constant. The intercept we measured is positive but statistically indistinguishable from zero across ten subjects. This acts as a null result rather than definitive proof of no contamination. Since every item in the corpus records its year, researchers can safely avoid this issue by restricting their comparisons exclusively to the 2026 exams, which were released on 19 June 2026.

\section{Ethics and Data Statement}

Every exam, key and score distribution we release is a published document of Vietnam's Ministry of
Education and Training, redistributed as published and cited per item. The candidate score records are
aggregate counts of marks and carry no personal identifier.

We publicly release the corpus, the ministry keys and the code that extracts and scores it at
\texttt{[PENDING: release URL]} under CC-BY-4.0, with the marking rule of Decision~764
implemented so that a later year's exams can be scored by the same command.

\section{Conclusion}

THPT-Ladder scores language models on 632 items from 21 official exams using Vietnam's published marking rules instead of flat accuracy. The ladder withholds 0.042 points a question from Qwen3.5-27B, a penalty that drops it thirteen percentile places on the 2025 History exam. It withholds 0.159 points from the weakest model, whose errors spread more thinly across questions. That statement accuracy and final marks are completely distinct quantities is a core property of the non-additive scheme. Even at a fixed statement accuracy, the mark the ladder pays spans a wide interval. With the strongest models earning full marks on eight of the eighteen exams, what separates them is exactly where their errors fall rather than how many they make. Convex partial credit is how an institution declines to pay for knowledge it cannot rely on. When a benchmark substitutes standard accuracy for these rules, it reports a competence the institution would never certify.

\end{document}